\documentclass[runningheads]{llncs}
\usepackage{booktabs}
\usepackage{multirow}
\usepackage{graphicx}
\usepackage{enumerate}
\usepackage{caption}
\usepackage{color}
\usepackage{graphicx}
\usepackage{amsmath}
\usepackage{amsfonts}
\usepackage{enumitem}
\usepackage{subcaption}

\begin{document}
\title{Revealing Reliable Signatures\protect\\by Learning Top-Rank Pairs}
\author{Xiaotong Ji\inst{1,\footnotemark[1]}\and
Yan Zheng\inst{1, \footnotemark[1]}\and
Daiki Suehiro\inst{1,2}\orcidID{0000-0001-8901-9063} \and
Seiichi Uchida\inst{1}\orcidID{0000-0001-8592-7566}}
\footnotetext[1]{The author contribute equally to this paper.}
\authorrunning{Ji. et al.}
\institute{Kyushu University, Fukuoka, Japan\\
\email{\{xiaotong.ji, yan.zheng\}@human.ait.kyushu-u.ac.jp}\\
\email{\{suehiro, uchida\}@ait.kyushu-u.ac.jp}\and
RIKEN Center for Advanced Intelligence Project, Tokyo, Japan}
\maketitle
\begin{abstract}
Signature verification, as a crucial practical documentation analysis task, has been continuously studied by researchers in machine learning and pattern recognition fields. In specific scenarios like confirming financial documents and legal instruments, ensuring the absolute reliability of signatures is of top priority. In this work, we proposed a new method to learn ``top-rank pairs'' for writer-independent offline signature verification tasks. By this scheme, it is possible to maximize the number of absolutely reliable signatures. More precisely, our method to learn top-rank pairs aims at pushing positive samples beyond negative samples, after pairing each of them with a genuine reference signature. In the experiment, BHSig-B and BHSig-H datasets are used for evaluation, on which the proposed model achieves overwhelming better pos@top (the ratio of absolute top positive samples to all of the positive samples) while showing encouraging performance on both Area Under the Curve (AUC) and accuracy.
\end{abstract}
\section{Introduction\label{sec:intro}}
%%% 1
As one of the most important topics in document processing systems, signature verification has become an indispensable issue in modern society~\cite{dey2017signet,zheng2019ranksvm}. Precisely, it plays important role in enhancing security and privacy in various fields, such as finance, medical, forensic agreements, etc. As innumerable significant documents are signed almost every moment throughout the world, automatically examining the genuineness of the signed signatures has become a crucial subject. Since misjudgment is hardly allowed especially in serious and formal situations like in forensic usages, obtaining ``highly reliable" signatures is of great importance.\par
%As a kind of biometric verification technology, signature verification has become an indispensable issue in modern society. Precisely, it plays important role in enhancing security and privacy in various fields, such as finance, security, forensic agreements, etc. As innumerable significant documents are signed almost every moment throughout the world, automatically examining the genuineness of the signed signatures has become a crucial subject. Since misjudgment is hardly allowed especially in serious and formal situations like in forensic usages, obtaining ``highly reliable" signatures is of great importance.\par

\begin{figure}[t!]
\begin{center}
\includegraphics[width=0.8\textwidth]{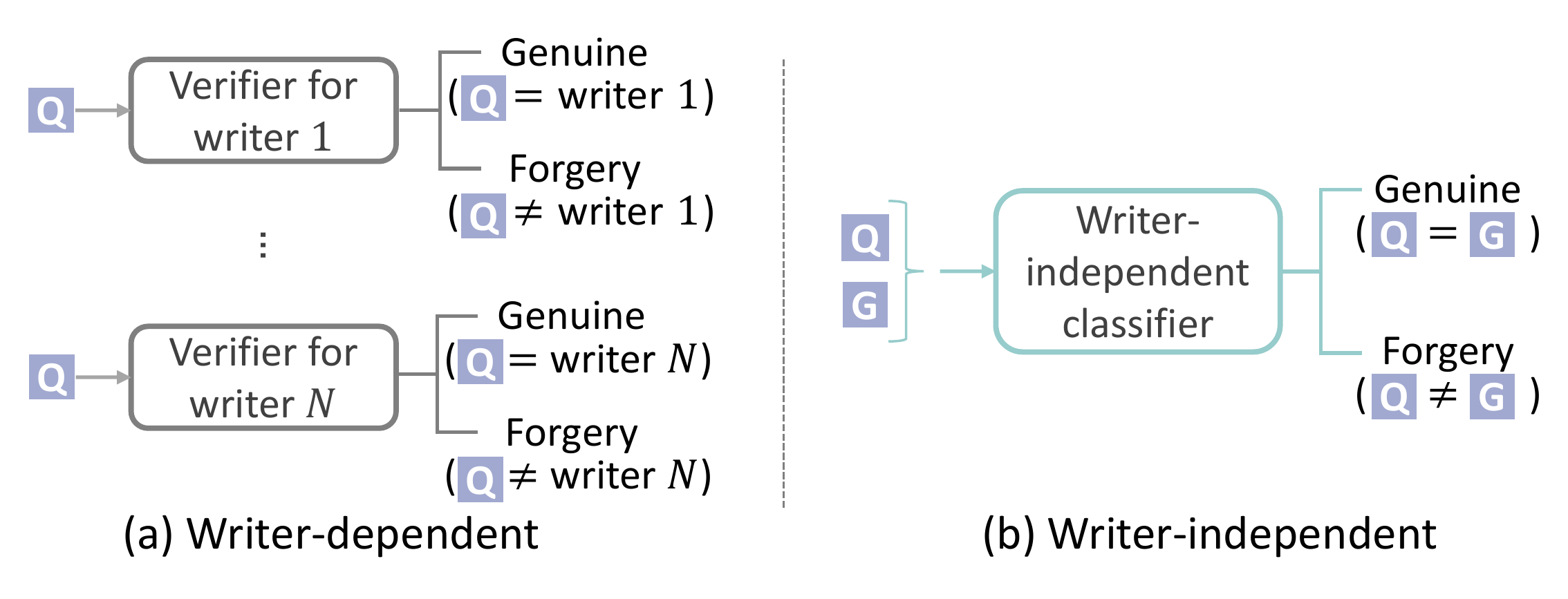}\\[-4mm]
\caption{Two scenarios of signature verification. `Q' and `G' are the query signature and a genuine reference  signature, respectively. }
\label{fig:dependentandindependent}
\end{center}
\medskip
\begin{center}
\includegraphics[width=\columnwidth]{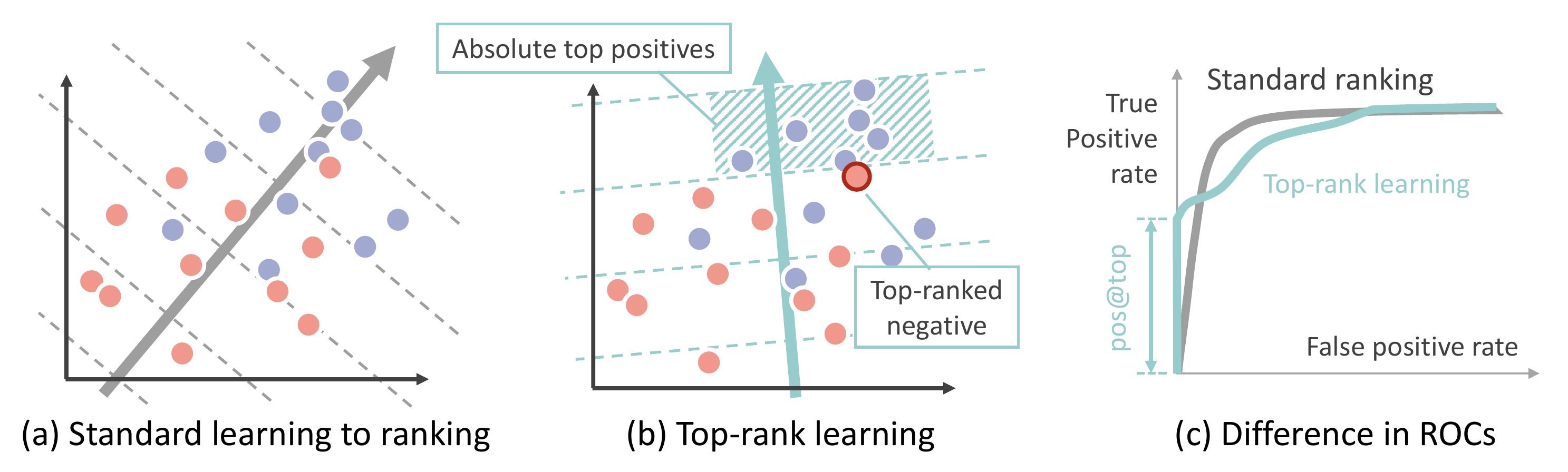}\\[-2mm]
\caption{(a), (b) Two ranking methods and (c)~their corresponding ROCs. In (a) and (b), a thick arrow is a ranking function that gives higher rank scores to positive samples (purple circles) than negative samples (pink circles). Each dotted line is an ``equi-distance'' line.}
% equi-distance → これにします！ という言い方はあるのよ．でもequal-scoreのほうがわかりやすいかな？？
\label{fig:rankingmethods}
\end{center}
\end{figure}

%Therefore, accurately and even absolutely verifying the authenticity of a given signature is of great importance, especially in serious and formal situations like in forensic usages.\par
%Signatures can diverse easily according to the mood or the external environment even signed by the same individual.
%our daily lives
%Unlike fingerprints that remain almost the same every time when recorded,
%%% 2
%As reviewed in Section 2, 
Fig.~\ref{fig:dependentandindependent} illustrates two scenarios of signature verification. In the writer-dependent scenario~(a), it is possible to prepare the verifiers specialized for individuals. In contrast, in the writer-independent scenario~(b), we can prepare only a single and universal classifier that judges whether a pair of a questioned signature (i.e., a query) and a genuine reference signature are written by the same person or not. Consequently, for a reliable writer-independent verification, the classifier needs (i)~to deal with various signatures of various individuals and (ii)~not to accept unreliable pairs with some confidence.\par

%%%%%%%%%%%%%%%%%%%%%%%%%%%%%%%%
\setlength{\belowcaptionskip}{-0.5cm}
%%%%%%%%%%%%%%%%%%%%%%%%%%%%%%%%%%%%
%
In this paper, we propose a new method to learn {\em top-rank pairs} for highly-reliable writer-independent signature verification. The proposed method is inspired by top-rank learning~\cite{li2014top,frery2017efficient,boyd2012accuracy}.
Top-rank learning is one of the ranking tasks but is different from the standard ranking task.
Fig.~\ref{fig:rankingmethods} shows the difference of top-rank learning from standard learning to rank. 
The objective of the standard ranking task~(a) is to determine the ranking function that evaluates positive samples much higher than negative samples as possible. This objective is equivalent to maximize AUC.  In contrast, top-rank learning~(b) has a different objective to maximize {\em absolute top positives},
which are highly-reliable positive samples in the sense that no negative sample has a higher rank than them.
In (c), the very beginning of the ROC of top-rank learning is a vertical part; this means that there are several positive samples that have no negative samples ranked higher than them. Consequently, Top-rank learning can derive absolute top positives as highly reliable positive samples. The ratio of absolute top positives over all positives is called ``{\em pos@top}.''  \par
The proposed method to learn top-rank pairs accepts a paired feature vector for utilizing the promising property of top-rank learning for writer-independent signature verification. As shown in Fig.~\ref{fig:dependentandindependent}~(b), the writer-independent scenario is based on the pairwise evaluation between a query and a genuine reference. To integrate this pairwise evaluation into top-rank learning, 
we first concatenate $q$ and $g$ into a single paired feature vector $\boldsymbol{x}=q\oplus g$, where $q$ and $g$ denote the feature vector of a query and a genuine reference, respectively. The paired vector $\boldsymbol{x}$ is treated as positive when $q$ is written by the genuine writer and denoted as $\boldsymbol{x}^+$; similarly, when $q$ is written by a forgery, $\boldsymbol{x}$ is negative and denoted as $\boldsymbol{x}^-$. \par
We train the ranking function $r$ with the positive paired vectors $\{\boldsymbol{x}^+\}$ and the negative paired vectors  $\{\boldsymbol{x}^-\}$, by top-rank learning. As the result, we could have highly reliable positives as absolute top positives; more specifically, a paired signature in the absolute top positives
is ``a more genuine pair'' than the most genuine-like negative pair (i.e., the most-skilled forgery), called {\em top-ranked negative}, therefore highly reliable. Although being an absolute top positive is much harder than just being ranked higher, we could make a highly reliable verification by using the absolute top positives and the trained ranking function $r$.\par
To prove the reliability of the proposed method in terms of pos@top, we conduct writer-independent offline signature verification experiments with two publicly-available datasets: BHSig-B and BHSig-H. We use SigNet~\cite{dey2017signet} 
as not only the extractor of deep feature vectors ($q$ and $g$) from individual signatures but also a comparative method. SigNet is the current state-of-the-art model of offline signature verification. The experimental results prove that the proposed method outperforms SigNet not only pos@top but also other conventional evaluation metrics, such as accuracy, AUC, FAR, and FRR.\par
Our contributions are arranged as follows:
\begin{itemize}[itemsep=0pt,topsep=5pt]
    \item We propose a novel method to learn top-rank pairs. This method is the first application of top-rank learning to conduct a writer-independent signature verification task to the best knowledge of the authors, notwithstanding that the concepts of top-rank learning and absolute top positives are particularly appropriate to highly reliable signature verification tasks.
    \item Experiments on two signature datasets have been done to evaluate the effect of the proposed method, including the comparison with the SigNet. Especially, the fact that the proposed method achieves higher pos@top proves that the trained ranking function gives a more reliable score that guarantees ``absolutely genuine'' signatures. 
\end{itemize}
\par

%%%%%%%%%%%%%%%%%%%%%%%%%%%%%%%%%%%%%%%
\section{Related Work}
%%% 1
The signature verification task has attracted great attention from researchers since it has been proposed~\cite{impedovo2008automatic,hafemann2017offline}. Generally, signature verification is divided into online~\cite{lee1996reliable} as well as offline~\cite{kalera2004offline,ferrer2012robustness} fashions. Online signatures offer pressure and stroke order information that is favorable to time series analysis methods~\cite{lai2018recurrent} like Dynamic Time Warping (DTW)~\cite{okawa2019template}. On the other hand, offline signature verification should be carried out only by making full use of image feature information~\cite{banerjee2021new}. As a result, acquiring efficacious features from offline signatures~\cite{hafemann2017offline,okawa2018bovw,ruiz2008offline} has become a highly anticipated challenge. \par

%%% 2.2
%Since offline signatures are rather ubiquitous but containing less usable information compared to online signatures, acquiring efficacious features from offline signatures~\cite{hafemann2017offline,okawa2018bovw} has become a highly anticipated challenge. Generally, hand craft geometric feature always play an important role in the process of offline signature verification tasks. Early trials on offline feature utilization always focus on distance measures as well as matching~\cite{kalera2004offline,ferrer2005offline,ruiz2008offline,abuhaiba2007offline}, bring about far-reaching influence to recent studies as well. Zois et al. put forward a grid-based feature matching model~\cite{zois2016offline} that shows great correlation with the ground truth, then successively proposed another scheme concentrates on sparse representation in signatures~\cite{zois2019comprehensive}. Moreover, Batool et al. in \cite{batool2020offline} has combined gray level co-occurrences matrix with geometric features using support vecter machines (SVM). Nevertheless, those beneficial geometric features show drawbacks to be vulnerable to keep effective when exposed to complicated backgrounds and noises.\par

%%% 2
In recent years, CNNs have been widely used in signature verification tasks thanks to their excellent representation learning ability ~\cite{hafemann2017learning,souza2018writer}. Among CNN-based models, Siamese network~\cite{melekhov2016siamese,guo2017learning} is one of the common choices when it comes to signature verification tasks. Specifically, a Siamese network is composed of two identically structured CNNs with shared weights, particularly powerful for similarity learning, which is a preferable learning objective in signature verification. For example, Dey et al. proposed a Siamese network-based model that optimizes the distances between two inputted signatures, shows outstanding performance on several famous signature datasets~\cite{dey2017signet}. Wei et al. also employed the Siamese network, and by utilizing inverse gray information with multiple attention modules, their work showed encouraging results as well~\cite{wei2019inverse}. However, none of the those approaches target revealing highly reliable genuine signatures. %, which are of great importance in real-world applications.

To acquire the highly reliable genuine signatures, learning to rank~\cite{trotman2005learning,burges2005learning} is a more reasonable approach than learning to classify. 
This is because learning to rank allows us to rank the signatures in order of genuineness.
%As for ranking methods, learning-to-rank has been widely studied especially in the field of relevance retrieval~\cite{joachims2002optimizing,trotman2005learning,burges2005learning,cao2007learning,valizadegan2009learning}. 
%The key point of learning-to-rank is to arrange queries in order in a supervised manner. 
Bipartite ranking~\cite{agarwal2005generalization} is one of the most standard learning-to rank-approach. 
The goal of the bipartite ranking is to find a scoring function that gives a higher value to positive samples than negative samples.
This goal corresponds to the maximization of AUC (Area under the ROC curve), and thus bipartite ranking has been used in various domains~\cite{usunier2011multiview,charoenphakdee2019learning,mehta2013efficient}.
%, as a special form of pairwise ranking approach, shows very interesting characteristics. By inputting a pair of samples into the same network in the training stage, bipartite ranking makes it possible to always rank a more related sample higher than a less related one, realizing outputting a ranking score for any new samples in the inference step. It is exactly because of its fascinating characteristic that some subsequent studies~\cite{kotlowski2011bipartite,agarwal2005stability,agarwal2005learnability,menon2016bipartite} has been done on it, while being widely applied in fields like documents retrieval~\cite{usunier2011multiview,charoenphakdee2019learning}, recommendation system~\cite{mehta2013efficient} and so on.\par

%%% 1.2
As a special form of bipartite ranking, the top-rank learning strategy~\cite{li2014top,frery2017efficient,boyd2012accuracy} possesses characteristics that are more suitable for absolute top positive hunting. 
In contrast to the standard bipartite ranking,
%Rather than 
%simply ranking positive samples higher than negative ones as the conventional bipartite ranking does (which is proved to be equal to maximize the AUC), 
top-rank learning aims at maximizing the absolute top positives, that is, maximizing the number of positive samples ranked higher than any negative sample. Therefore, top-rank learning is suitable for
some tough tasks that require possibly highly reliable (e.g., medical diagnosis~\cite{zheng2021top}). 
To the best of our knowledge, this is the first application of top-rank learning to the signature verification task.
%serious applications such as private property and personal safety, as in scenarios like signature verification and medical diagnosis.\par

%%% 1.3

TopRank CNN~\cite{zheng2021top} is a representation learning version of the conventional top-rank learning scheme, which combines the favorable characteristics of both CNN and the top-rank learning scheme. To be more specific, when encountered with entangled features in which positive is chaotically tied with negative, conventional top-rank learning methods without representation learning capability like TopPush~\cite{li2014top} can hardly achieve a high pos@top. That is to say, the representation learning ability of CNN structure makes the TopRank CNN a more powerful top-rank learning method. Moreover, for avoiding the easily-happen over-fitting phenomenon, TopRank CNN considerately attached the max operation with the p-norm relaxation.

%%% 4
Despite the superiority of ranking schemes, studies that apply ranking strategies on signature verification tasks are still in great demand. Chhabra et al. in~\cite{chhabra2019siamese} proposed a Deep Triplet Ranking CNN, aiming at ranking the input signatures in genuine-like order by minimizing the distance between genuine and anchors. In the same year, Zheng et al. proposed to utilize RankSVM for writer-dependent classification, to ensure the generalization performance on imbalanced data~\cite{zheng2019ranksvm}. However, even if they care about ranking results to some extent, no existing studies have been dedicated to the absolute top genuine signatures yet. To address this issue, this work is mainly designed to focus on obtaining the absolute top genuine signatures, implemented by learning top-rank pairs for writer-independent offline signature verification tasks.

\setlength{\belowcaptionskip}{-0.5cm}
\begin{figure}[t!]
\begin{center}
\includegraphics[width=\columnwidth]{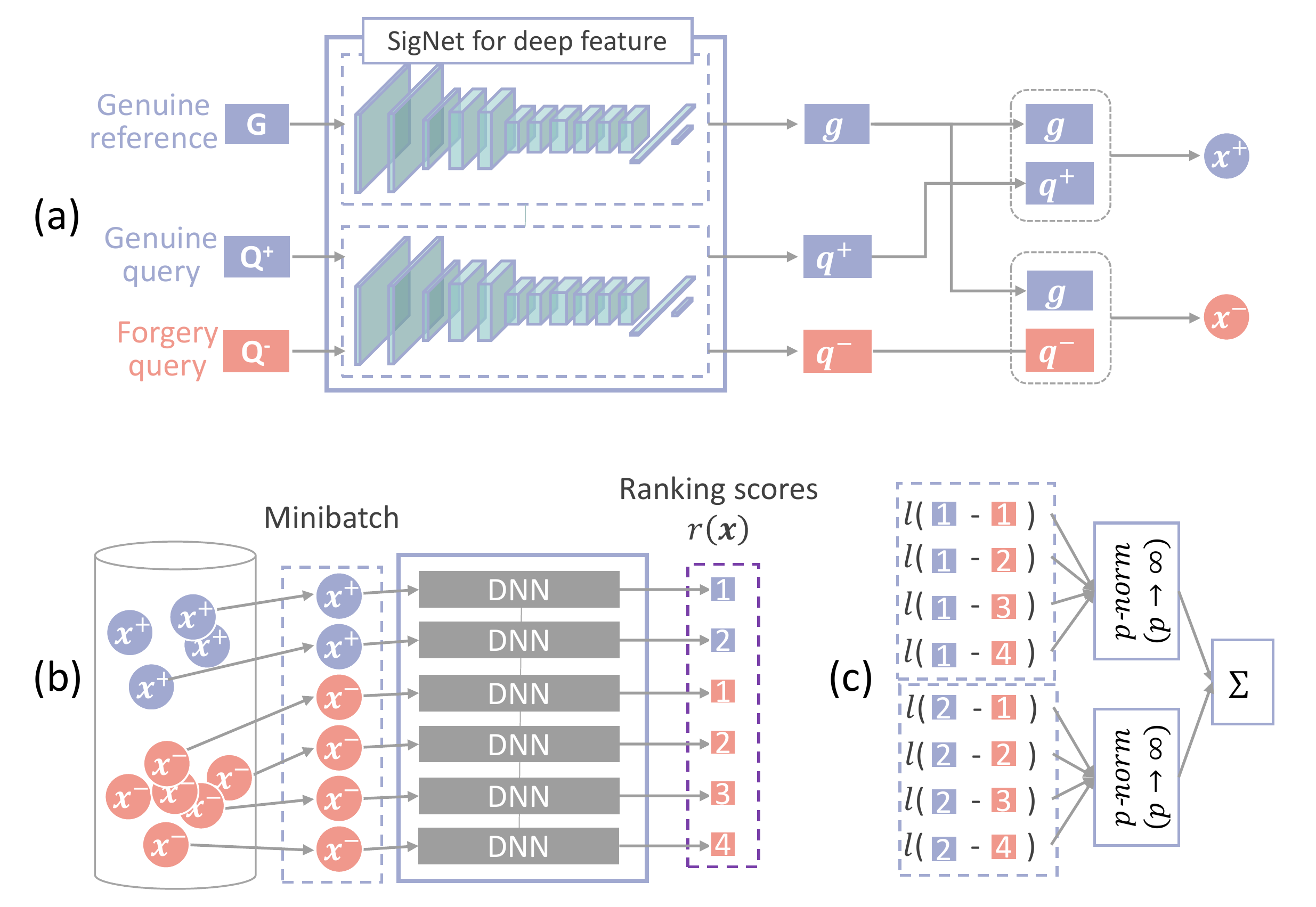}\\[-2mm]
\caption{The overall structure of the proposed method to learn {\em top-rank pairs}  for writer-independent signature verification. (a)~Feature representation of paired samples.  (b)~Learning top-rank pairs with their representation. (c)~Top-rank loss function $\mathcal{L}_{\mathrm{TopRank}}$.}
\label{fig:overall}
\end{center}
\end{figure}
%%%%%%%%%%%%%%%%%%%%%%%%%%%%%%%%%%%%
\section{Learning Top-Rank Pairs}
%%%%%%%%%%%%%%%%%%%%%%%%%%%%%%%%%%%%%%%%%%%
Fig.~\ref{fig:overall} shows the overview of the proposed method to learn top-rank pairs for writer-independent offline signature verification. The proposed method consists of two steps: a representation learning step and a top-rank learning step. Fig.~\ref{fig:overall}~(a) shows the former step and (b) and (c) show the latter step.
%-----------------------------------
\subsection{Feature representation of paired samples\label{sec:concat-feature}}
As shown in Fig.~\ref{fig:dependentandindependent}~(b), each input is a pair of a genuine reference sample $g$ and a query sample $q$ for writer-independent signature verification. Then the paired samples $(g,q)$ are fed to some function to evaluate their discrepancy. If the evaluation result shows a large discrepancy, the query is supposed to be a forgery; otherwise, the query is genuine.\par
Now we concatenate the two $d$-dimensional feature vectors ($g$ and $q$) into a $2d$-dimensional single vector as shown in Fig.~\ref{fig:overall}~(a). Although the concatenation doubles the feature dimensionality, it allows us to treat the paired samples in a simple way. Specifically, we consider a (Genuine $g$, Genuine $q^+$)-pair as a positive sample with the feature vector $\boldsymbol{x}^+=g\oplus q^+$ and a (Genuine $g$, Forgery $q^-$)-pair as a negative sample with $\boldsymbol{x}^-=g\oplus q^-$. If we have $m$ (Genuine, Genuine)-pairs and $n$ (Genuine, Forgery)-pairs, we have two sets $\mathbf{\Omega}^+ = \{\boldsymbol{x}_i^+\ |\ i=1,\ldots,m\}$ and  $\mathbf{\Omega}^-=\{\boldsymbol{x}_j^+\ |\ j=1,\ldots,n\}$. \par
Under this representation, the writer-independent signature verification task is simply formulated as a problem to have a function $r(\boldsymbol{x})$ that gives a large value for 
$\boldsymbol{x}_i^+$ and a small value for $\boldsymbol{x}_j^-$. Ideally, we want to have $r(\boldsymbol{x})$ that satisfies $r(\boldsymbol{x}_i^+) > r(\boldsymbol{x}_j^-)$ for arbitrary $\boldsymbol{x}_i^+$ and $\boldsymbol{x}_j^-$. In this case, we have a constant threshold $\theta$ that satisfies $\max_j r(\boldsymbol{x}_j^-)< \theta < \min_i r(\boldsymbol{x}_i^+)$. If $r(\boldsymbol{x})> \theta$, $\boldsymbol{x}$ is simply determined as a (Genuine, Genuine)-pair. However, in reality, we do not have the ideal $r$ in advance; therefore we need to optimize (i.e., train) $r$ so that it becomes closer to the ideal case under some criterion. In \ref{sec:optimize}, pos@top is used as the criterion so that trained $r$ gives more absolute tops.\par
As indicated in Fig.~\ref{fig:overall}~(a), each signature is initially represented as a $d$-dimensional vector $g$ (or $q$) by SigNet~\cite{dey2017signet}, which is still a state-of-the-art signature verification model realized by metric learning with the contrastive loss.  Although it is possible to use another model to have the initial feature vector, we use SigNet throughout this paper. The details of SigNet will be described in \ref{sec:signet}.

%-----------------------------------
\subsection{Optimization to learn top-rank pairs\label{sec:optimize}}
We then use a top-rank learning model for optimizing the ranking function $r(\boldsymbol{x})$. As noted in Section~\ref{sec:intro}, top-rank learning aims to maximize pos@top, which is formulated as:
\begin{align}
\label{eq:posatop}
    \mathrm{pos@top}=\frac{1}{m}\sum_{i=1}^m 
    I\left(r(\boldsymbol{x}_i^+) > 
    \max_{1\leq j\leq n}r(\boldsymbol{x}_j^-)
    \right),
\end{align}
where $I(z)$ is the indicator function. pos@top evaluates the number of positive samples with a higher value than any negative samples. The positive samples that satisfy the condition in Eq.\ref{eq:posatop} are called absolute top positives or just simply absolute tops. 
Absolute tops are very ``reliable'' positive samples because they are more positive than 
the top-ranked negative, that is, the ``hardest'' negative $\max_{1\leq j\leq n}r(\boldsymbol{x}_j^-).$ \par
Among various optimization criteria, pos@top has promising properties for the writer-independent signature verification task. Maximization of pos@top is equivalent to the maximization of absolute tops --- this means we can have reliable positive samples to the utmost extent. In a very strict signature verification task, the query sample $q$ is verified as genuine only when the concatenated vector $\boldsymbol{x}=g\oplus q$ becomes one of the absolute tops. Therefore, having more absolute tops by maximizing pos@top will give more chance that the query sample is completely trusted as genuine. \par
Note that we call $r(\boldsymbol{x})$ as a ``ranking'' function, instead of just a scoring function. In Eq.~(\ref{eq:posatop}), the value of the function $r$ is used just for the comparison of samples. This suggests that the value of $r$ has no absolute meaning. In fact, if we have a maximum pos@top by a certain $r(\boldsymbol{x})$, $\phi(r(\boldsymbol{x}))$ also achieves the same pos@top, where $\phi$ is a monotonically-increasing function. Consequently, the ranking function $r$ specifies the order (i.e., the rank) among samples.
\par
We will optimize $r$ to maximize $\boldsymbol{x}^+$ in pos@top. Top-Rank Learning is the existing problem to maximize pos@top for a training set whose samples are individual (i.e., unpaired) vectors.  In contrast, our problem to learn top-rank pairs is a new ranking problem for the paired samples, and applicable to various ranking problems where the relative relations between two vectors are important~\footnote{Theoretically, learning top-rank pairs can be extended to handle vectors obtained by concatenating three or more individual vectors. With this extension, our method can rank the mutual relationship among multiple vectors.}.\par
\par
As noted in \ref{sec:DNN} and shown in Fig.~\ref{fig:overall}~(b), we train $r$ along with a deep neural network (DNN) to have a reasonable feature space to have more pos@top. However, there are some risks when we maximize Eq.~(\ref{eq:posatop}) directly using a DNN, if it has a high representation flexibility.
The most realistic case is that the DNN overfits some outliers or noise. For example, if a negative outlier is distributed over the positive training samples, achieving the perfect pos@top is not a reasonable goal.\par
To avoid such risks, we employ the $p$-norm relaxation technique~\cite{JMLR:v10:rudin09b,zheng2021top}.
More specifically, we convert the maximization of pos@top into the minimization of the following loss:
\begin{align}
\label{eq:toprankloss}
    \mathcal{L}_{\mathrm{TopRank}}\left(\mathbf{\Omega^+},\mathbf{\Omega^-}\right)=
    \frac{1}{m}\sum_{i=1}^m\left(
    \sum_{j=1}^n\left(
    l(r(\boldsymbol{x}_i^+) - 
     r(\boldsymbol{x}_j^-)
    )\right)^p\right)^{\frac{1}{p}},
\end{align}
where $l(z) = \log(1+e^{-z})$ is a surrogate loss. Fig.~\ref{fig:overall}~(c) illustrates $\mathcal{L}_{\mathrm{TopRank}}$.
When $p=\infty$, Eq.~(\ref{eq:toprankloss}) is reduced to $\mathcal{L}_{\mathrm{TopRank}}= \frac{1}{m}\sum_{i=1}^m \max_{1\leq j \leq n} l(r(\boldsymbol{x}_i^+) - r(\boldsymbol{x}_j^-))$, which is equivalent to the original pos@top loss of Eq.~(\ref{eq:posatop}). If we set $p$ at a large value (e.g., 32), the Eq.~(\ref{eq:toprankloss}) approaches the original loss of pos@top.
In \cite{zheng2021top}, it is noted that it is better not to select 
a too large $p$, because it has a risk of over-fitting and the 
overflow error in the implementation.
% Uchida 2:40
\par
\setlength{\belowcaptionskip}{-0.5cm}
\begin{figure}[t!]
\begin{center}
\includegraphics[width=\columnwidth]{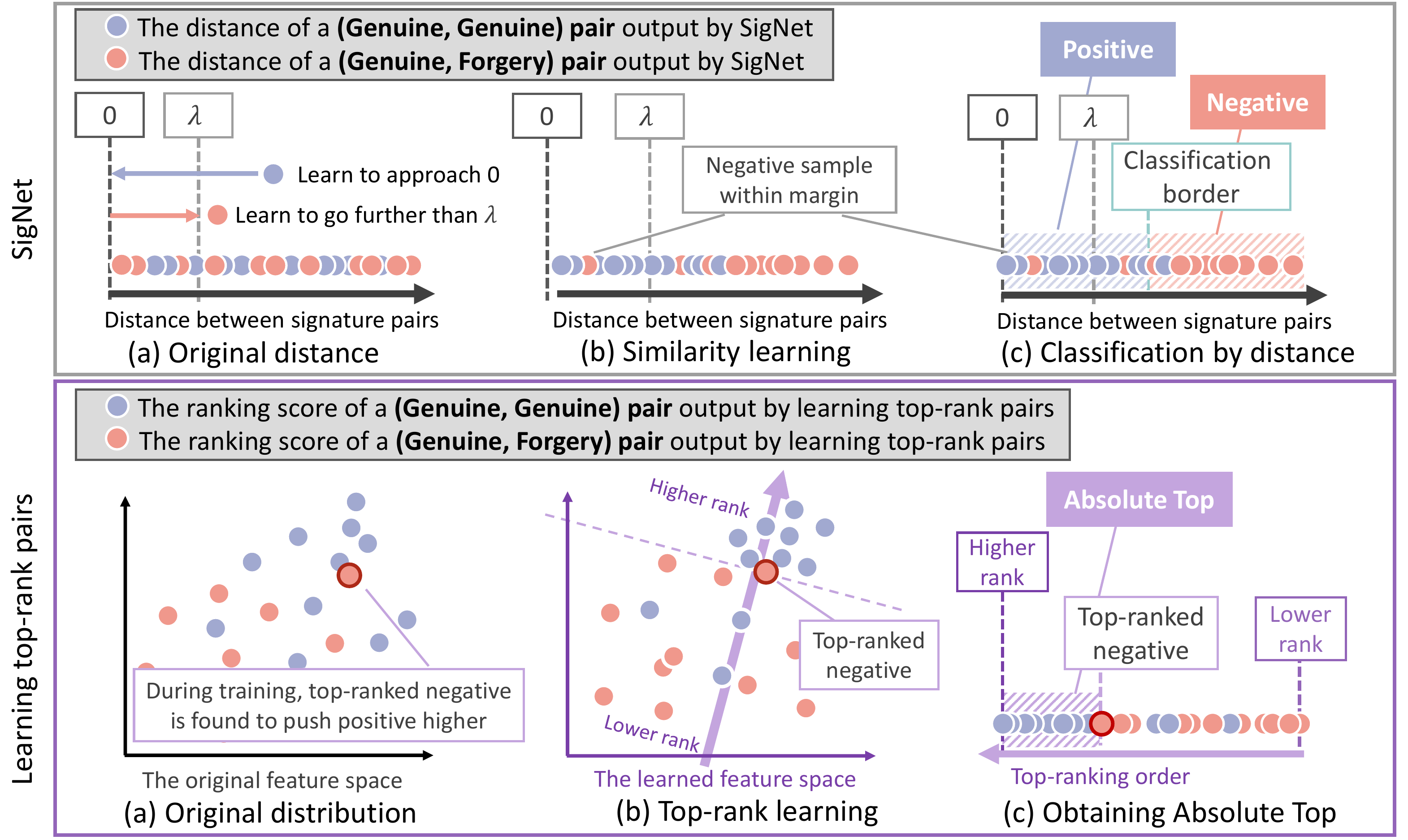}\\[-2mm]
\caption{The learning mechanisms of SigNet (top) and the learning top-rank pairs (bottom).}
\label{fig:signettoprank}
\end{center}
\end{figure}
%-----------------------------------
\subsection{Learning top-rank pairs with their representation\label{sec:DNN}}
In order to have a final feature representation to have a more pos@top, we apply a DNN to convert  $\boldsymbol{x}$ non-linearly during the training process of $r$. Fig.~\ref{fig:overall}~(b) shows 
the process with the DNN. Each of the paired feature vectors in a minibatch is fed to a DNN. In the DNN, the vectors are converted to another feature space and then their ranking score $r$ is calculated. The parameters of DNN are trained by the loss function $\mathcal{L}_{\mathrm{TopRank}}$.  
%-------------------------------------------------
\subsection{Initial features by SigNet\label{sec:signet}}
As noted in \ref{sec:concat-feature}, we need to have initial vectors ($g$ and $q$) for individual signatures by an arbitrary signature image representation method. 
SigNet~\cite{dey2017signet} is the current state-of-the-art signature verification 
method and achieved high performance in standard accuracy measures.
As shown in Fig.~\ref{fig:overall}~(a), SigNet is based on metric learning with a contrastive loss and takes a pair of a reference signature and a query signature as its input images. For all pairs of reference and query signatures, SigNet is optimized to decrease the distance between (Genuine, Genuine)-pairs and increase the distance between (Genuine, Forgery)-pairs. The trained network can convert a reference image into $g$ 
and a query into $q$. Then a positive sample $\boldsymbol{x}_i^+$ or a negative sample $\boldsymbol{x}_j^-$ is obtained by the concatenation $g\oplus q$, as described in  \ref{sec:concat-feature}. \par
Theoretically, we can conduct the end-to-end training of SigNet in Fig.~\ref{fig:overall}~(a) and DNN in (b). In this paper, however, we fix the SigNet model after its independent training with the contrastive loss. This is simply to make the comparison between the proposed method and SigNet as fair as possible. (In other words, we want to purely observe the effect of pos@top maximization and thus minimize the extra effect of further representation learning in SigNet.)
\par
One might misunderstand that the metric learning result by SigNet and the ranking result by top-rank learning in the proposed method are almost identical; however, as shown in Fig.~\ref{fig:signettoprank}, they are very different. As we emphasized so far, the proposed method aims to have more pos@top; this means we have a clear boundary between the absolute tops and the others. In contrast, SigNet has no such function. Consequently, SigNet might have a risk that a forgery has a very small distance with a genuine. Finally, this forgery will be wrongly considered as one of the reliable positives, which are determined by applying a threshold $\lambda$ to the distance by SigNet.

\section{Experiments}
In this section, we demonstrate the effectiveness of the proposed method on signature verification tasks. Specifically, we consider a comparative experiment with SigNet which is known as the outstanding method for the signature verification tasks. %, which is also involved as the feature extraction apart of the proposed model.

\subsection{Datasets}
In this work, the BHSig260 offline signature dataset\footnote{Available at http://www.gpds.ulpgc.es/download} is used for the experiments\footnote{CEDAR dataset that is also used in~\cite{dey2017signet} was not used in this work because it has achieved 100\% accuracy in the test set. Besides, GPDS 300 and GPDS Synthetic Signature Corpus datasets are restricted from obtaining}, which composes two subsets where one set of signatures are signed in Bengali (named BHSig-B) and the other in Hindi (named BHSig-H). The BHSig-B dataset includes 100 writers in total, each of them possesses 24 genuine signatures and 30 skillfully forged signatures. On the other hand, BHSig-H dataset contains 160 writers, each of them own genuine and forged signatures same as BHSig-B. In the experiments, both of the datasets are divided into training, validation, and test set to the ratio of 8: 1: 1.\par

Following the writer-independent setting, 
we evaluate the verification performance using the pair of signatures.
That is, the task is to verify that the given pair is (Genuine, Genuine) or (Genuine, Forgery).
We prepare a total of 276 (Genuine, Genuine)-pairs for each writer
and a total of 720 (Genuine, Forgery)-pairs for each writer.
%the performance of models is evaluated in a pair-wise manner. Therefore, we teamed up the signatures into pairs in advance for these two datasets in the same way. Specifically, to make up the positive inputs, each genuine signature gets teamed with another genuine signature, leading to 276 (Genuine, Genuine)-pairs for each writer. As for the negative inputs, each forged signature gets teamed with every genuine signature as well but not with themselves, resulting in $24\times 30=720$ (Genuine, Forgery)-pairs for each writer.

\subsection{Experimental Settings}
%%%%%%%%%%%%%%%%%%%%%%%%%%%%%%%%%%%%%%%%%%
\subsubsection{Setting of the SigNet}
\label{subsec:setting_signet}
SigNet is also based on a Siamese network architecture, whose optimization objective is similarity measurement. In the experiment, we followed the training setting noted in~\cite{dey2017signet}, except for the modifications of the data division. %After training, the performance of the learned SigNet could be evaluated. Besides, as mentioned in Section 3.1, we also extract features of signatures from its hidden layer to compose discriminative inputs for top-rank learning (see Fig.~\ref{fig:overall} for details).\par

%In the process of evaluation, signature pairs should be sorted according to the scores assigned by the SigNet. Since SigNet is trained to optimize the distance between signatures within a pair (see Equation~\ref{eq:SigNetloss}), the shorter the distance, the more similar the pair, and vice versa. Therefore, when sorting a set of signature pairs by SigNet in a similar to dissimilar order, their corresponding scores will be organized in ascending order.\par
 
\subsubsection{Setting of the proposed method}
As introduced in Section~\ref{sec:signet}, we used the extracted features from the trained SigNet (124G Floating point operations (FLOPs)) as the initial features.
For the learning top-rank pairs with their representation, we used a simple architecture, 4 fully-connected layers (2048, 1024, 512 and 128 nodes respectively) with ReLU function  (1G FLOPs).
%The main body of the proposed TopRank CNN is also a Siamese architecture, who contains a pair of identically structured weight-sharing CNNs.
%The architecture of TopRank CNN in Fig.~\ref{fig:signettoprank} is constructed by DNNs. Each DNN is formed by four fully connected layers (each attached by a rectified linear unit (ReLU)).
%Additionally, there is a unique yet interesting technique in the training process of TopRank CNN, which is to organize training samples into mini-batches. Whenever a mini-batch is made up, a top-ranked negative is found to push positive higher by the p-norm relaxation, suggesting a must that both positive and negative (better to be more than positive) should be involved in each mini-batch, 
%as shown in the lower middle part of Fig.~\ref{fig:overall}. In our case, the mini-batches are composed of 5 (Genuine, Genuine)-pairs and 40 (Genuine, Forgery)-pairs.\par
%local 
%three convolutional layers
%Given the definition of loss for TopRank CNN in Equation~\ref{eq:toprankloss} with a non-increasing surrogate loss $l(z)$, it suggests that the larger the $r(x)$, the more similar the signatures are within the pair. Thus, the order of signature pairs is sorted within the descending rule of their ranking scores.
The hyper-parameter $p$ of the loss function~\ref{eq:toprankloss} is chosen from $\{2, 4, 8, 16, 32\}$ based on the validation pos@top. 
As a result, we obtained $p$=4 for BHSig-B and $p$=16 for BHSig-H. The following results and the visualization are obtained by using these hyper-parameters.
%Among the choices of $p$, the smaller it is, the milder learning effect it incurs. As the result, considering the numerical overflow, we fixed the maximum value of $p$ to 32.\par
%Although $p$ could be set to infinitely large, no obvious change happened when it was set to be larger than 32 as we have tested. 

\subsection{Evaluation metrics}
%Equal Error Rate (EER)? 
In the experiment, pos@top, accuracy, AUC, False Rejection Rate (FRR), and False Acceptance Rate (FAR) are used to comprehensively evaluate the proposed  method and SigNet. 
\begin{itemize}[itemsep=0pt,topsep=5pt]
\item \textbf{pos@top}: The ratio of the absolute top (Genuine, Genuine) signature pairs to the number of all of the (Genuine, Genuine) signature pairs (see also Eq.~(\ref{eq:posatop})).
\item Accuracy: The maximum result of the average between True Positive Rate (TPR) and True Negative Rate (TNR), following the definition in~\cite{dey2017signet}.
\item AUC: Area under the ROC curve.
\item FAR: The ratio of the number of falsely accepted (Genuine, Forgery) signature pairs divided by the number of all (Genuine, Forgery) signature pairs.
\item FRR: The ratio of the number of falsely rejected (Genuine, Genuine) signature pairs divided by the number of all (Genuine, Genuine) signature pairs.
\end{itemize}
%It is worth noting that the meaning of outputs of TopRank CNN and the outputs of SigNet are different. That is, for example, given an extremely similar (Genuine, Genuine) pair, TopRank CNN would give it a very high score, while SigNet would assign it with a very low score. Consequently, in the evaluation stage, scores from TopRank CNN are evaluated in descending order, and scores from SigNet in ascending order.
\subsection{Quantitative and qualitative evaluations}

%%% table to compare AUC accuracy and pos@top
%%%%%%%%%%%%%%%%%%%%%%%%%%%%%%%%%%%%%%%%
\begin{table}[t!]
\centering
\caption{The comparison between the proposed method and SigNet on BHSig-B and BHSig-H datasets.}\label{tab1}
\setlength{\tabcolsep}{1mm}
\begin{tabular}{ccccccc}
\toprule
Dataset & Approaches & {\bfseries pos@top ($\uparrow$)} & Accuracy ($\uparrow$) & AUC ($\uparrow$)& FAR ($\downarrow$)& FRR ($\downarrow$)\\
\hline
\specialrule{0em}{1pt}{1pt}
\multirow{2}{*}{BHSig-B} & {\bfseries proposed } & {\bfseries 0.283} & {\bfseries 0.806} & {\bfseries 0.889} & {\bfseries 0.222} & {\bfseries0.222} \\
& SigNet & 0.000 & 0.756 & 0.847 & 0.246 & 0.247 \\
\hline
\specialrule{0em}{1pt}{1pt}
\multirow{2}{*}{BHSig-H} & {\bfseries proposed } & {\bfseries 0.114} & {\bfseries 0.836} & {\bfseries 0.908} & {\bfseries 0.179} & {\bfseries 0.178} \\
& SigNet & 0.000 & 0.817 & 0.891 & 0.192 & 0.192 \\
\bottomrule
\end{tabular}
\end{table}
%%%%%%%%%%%%%%%%%%%%%%%%%%%%%%%%%%%%%%%%%%%%%%
%%% fig ROC
%%%%%%%%%%%%%%%%%%%%%%%%%%%%%%%%%%%%%%%%%%%
\setlength{\belowcaptionskip}{-0.1cm}
\begin{figure}[t!]
\centering
\begin{subfigure}{0.495\columnwidth}
\centering
\includegraphics[width=6cm]{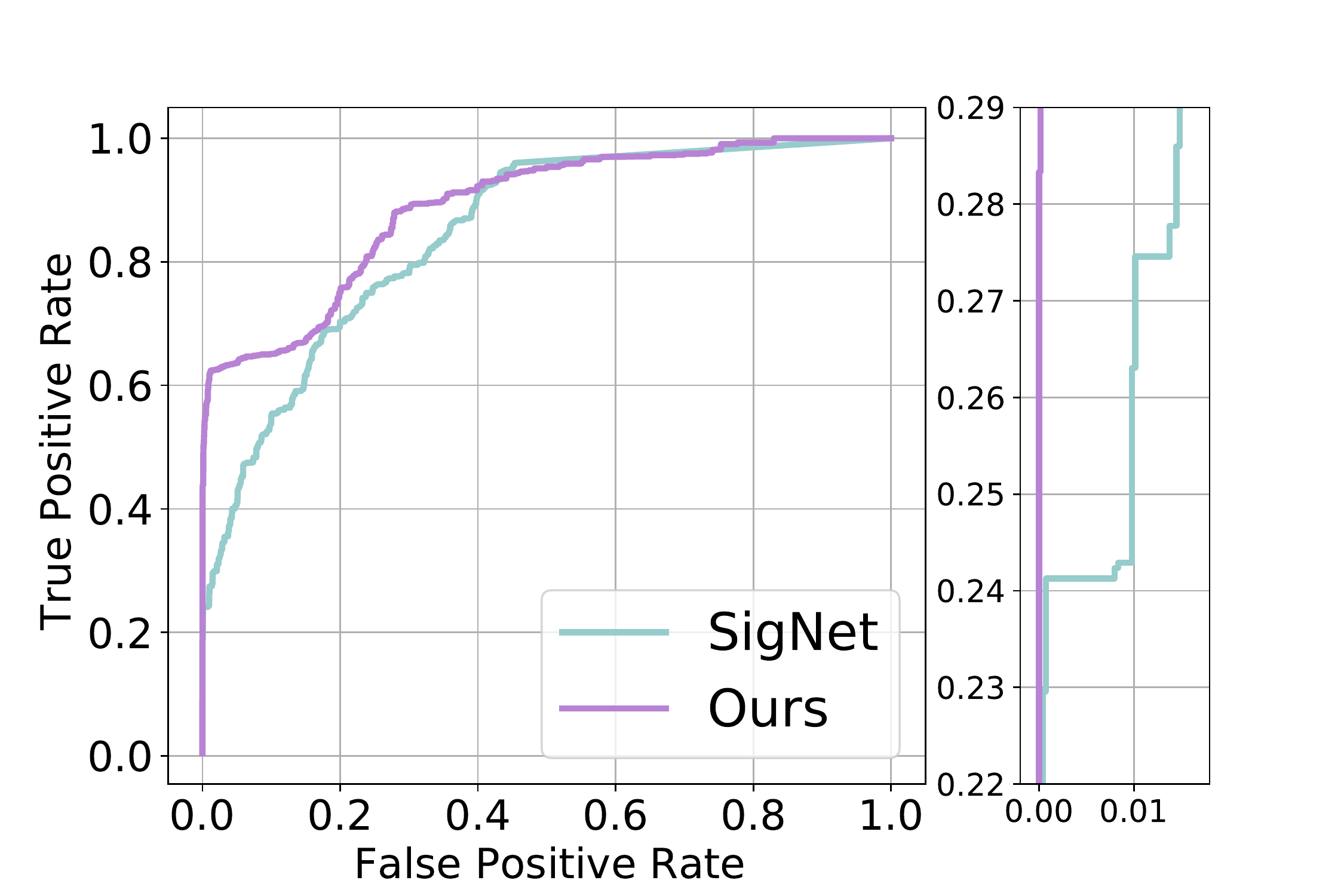}
\caption{ROC curves (full and zoom) on BHSig-B}
\label{chutian3}%文中引用该图片代号
\end{subfigure}
\centering
\begin{subfigure}{0.495\columnwidth}
\centering
\includegraphics[width=6cm]{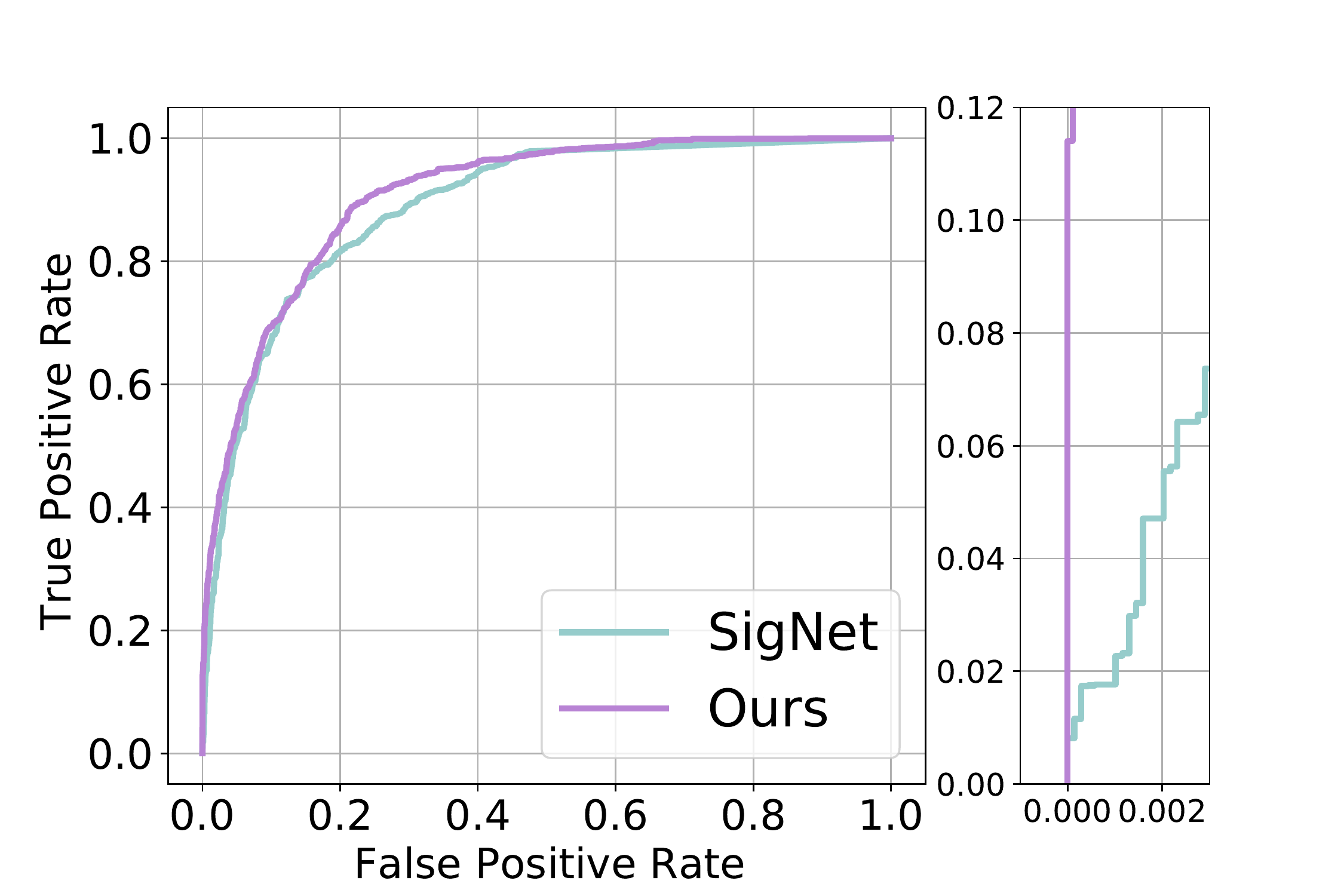}
\caption{ROC curves (full and zoom) on BHSig-H}
\label{chutian3}%文中引用该图片代号
\end{subfigure}
\caption{The comparison of ROC curves between the proposed method and SigNet on BHSig-B and BHSig-H datasets.}
\label{fig:roc}
\end{figure}
%%%%%%%%%%%%%%%%%%%%%%%%%%%%%%%%%%%%%%%

%%% fig PCA
%%%%%%%%%%%%%%%%%%%%%%%%%%%%%%%%%%%%%%%%%%%
\setlength{\belowcaptionskip}{-0.0cm}
\begin{figure}[t!]
\centering
\begin{subfigure}{0.495\columnwidth}
\centering
\includegraphics[width=6cm]{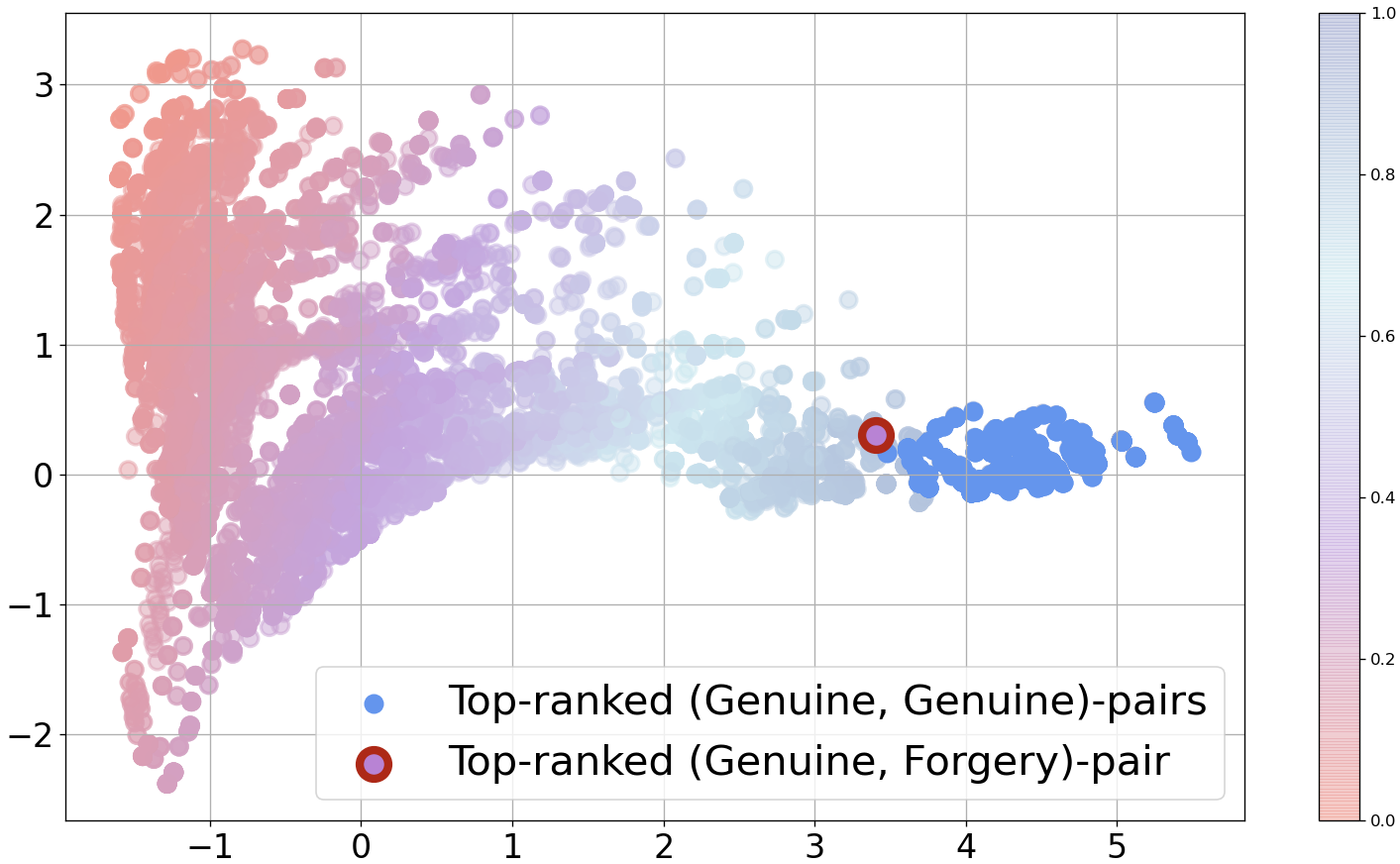}
\caption{PCA result of the proposed method on BHSig-B}
\label{fig:bengaliPCA}%文中引用该图片代号
\end{subfigure}
\centering
\begin{subfigure}{0.495\columnwidth}
\centering
\includegraphics[width=6cm]{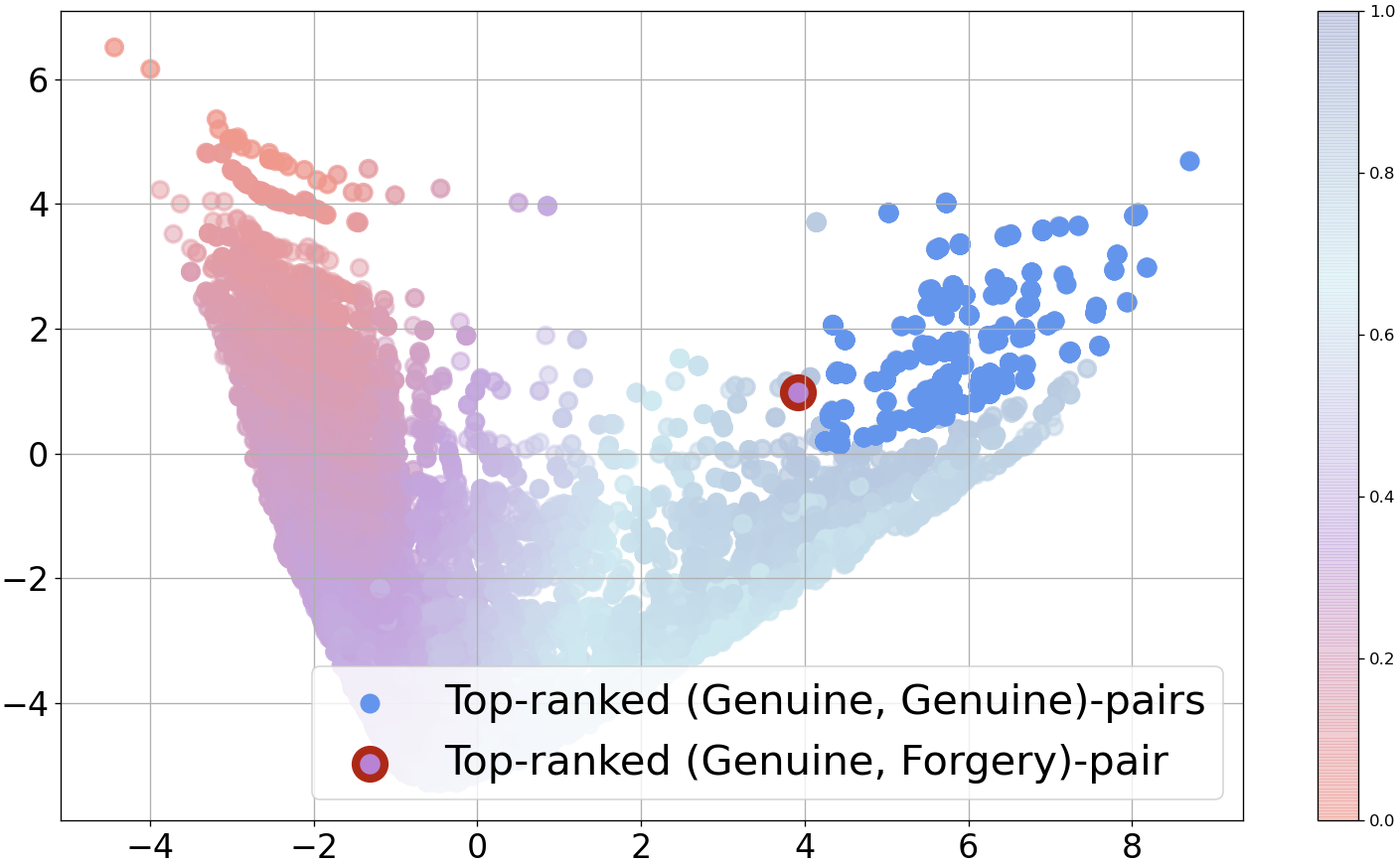}
\caption{PCA result of the proposed method on BHSig-H}
\label{fig:hindiPCA}%文中引用该图片代号
\end{subfigure}\\
\medskip
\centering
\begin{subfigure}{0.495\columnwidth}
\centering
\includegraphics[width=6cm]{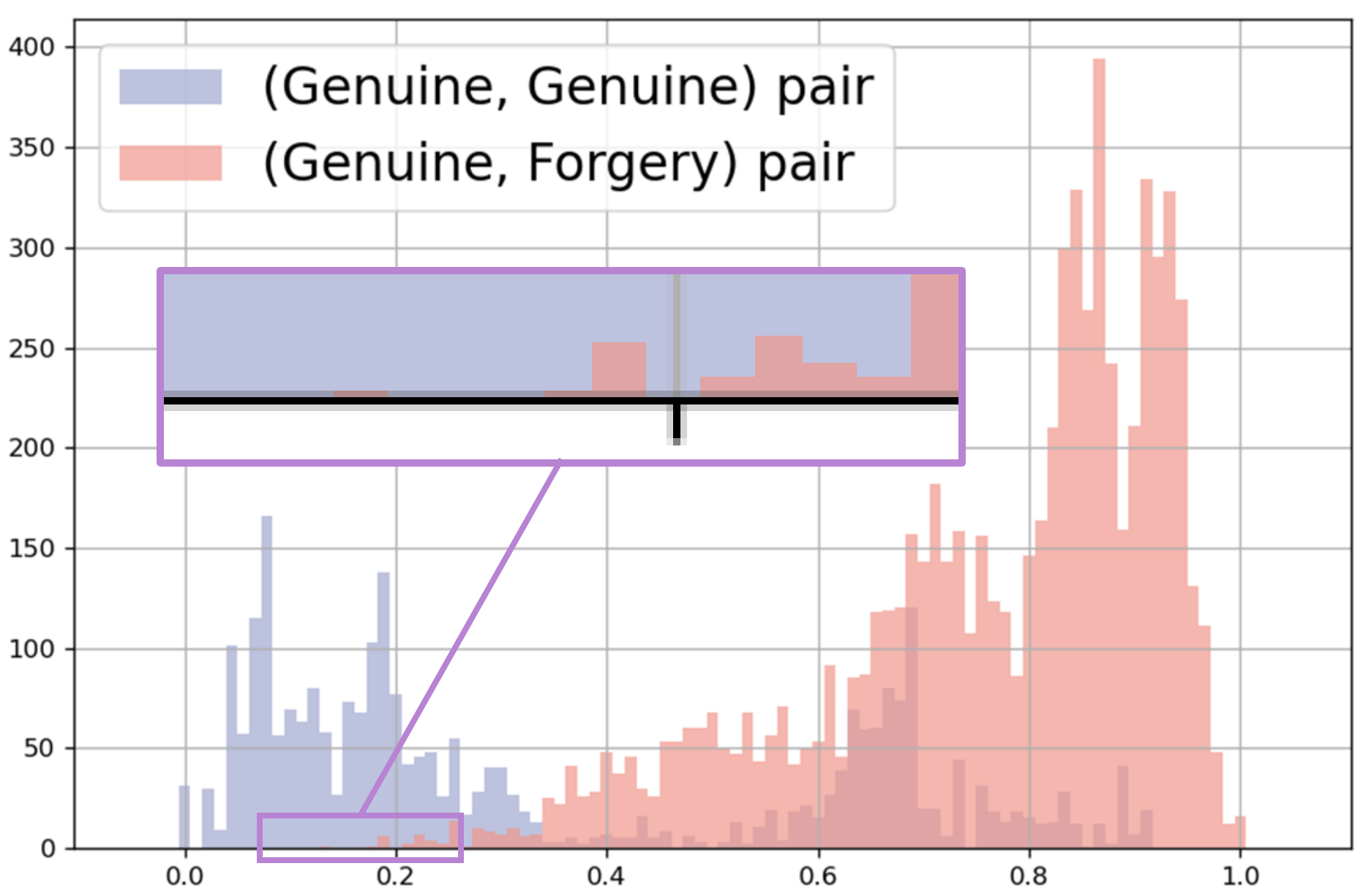}
\captionsetup{justification=centering}
\caption{Histogram of the ranking scores of the proposed method with a zoomed view on BHSig-B}
\label{fig:bengalitophist}%文中引用该图片代号
\end{subfigure}
\centering
\begin{subfigure}{0.495\columnwidth}
\centering
\includegraphics[width=6cm]{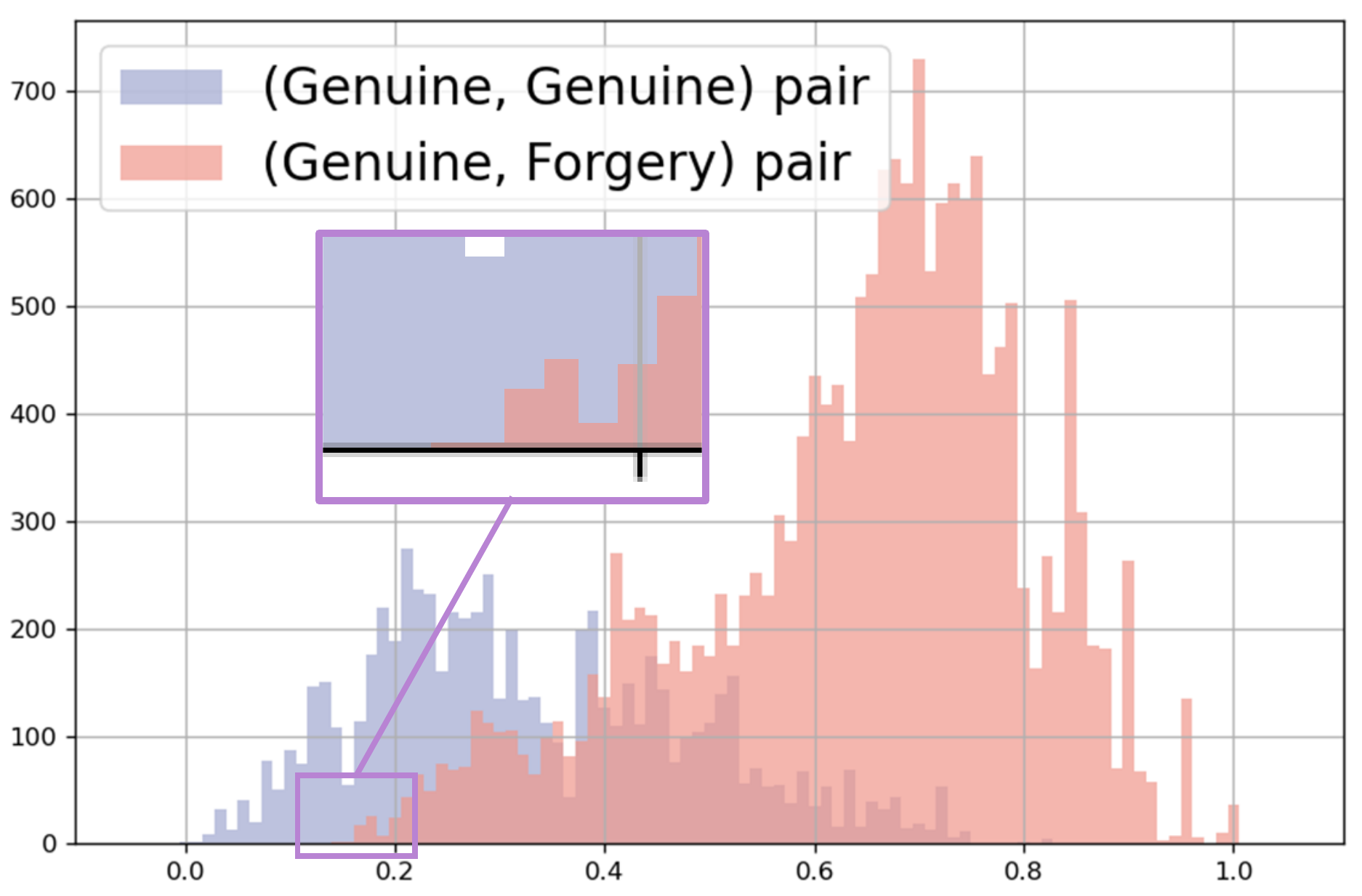}
\captionsetup{justification=centering}
\caption{Histogram of the ranking scores of the proposed method with a zoomed view on BHSig-H}
\label{fig:hinditophist}%文中引用该图片代号
\end{subfigure}\\
\medskip
\centering
\begin{subfigure}{0.495\columnwidth}
\centering
\includegraphics[width=6cm]{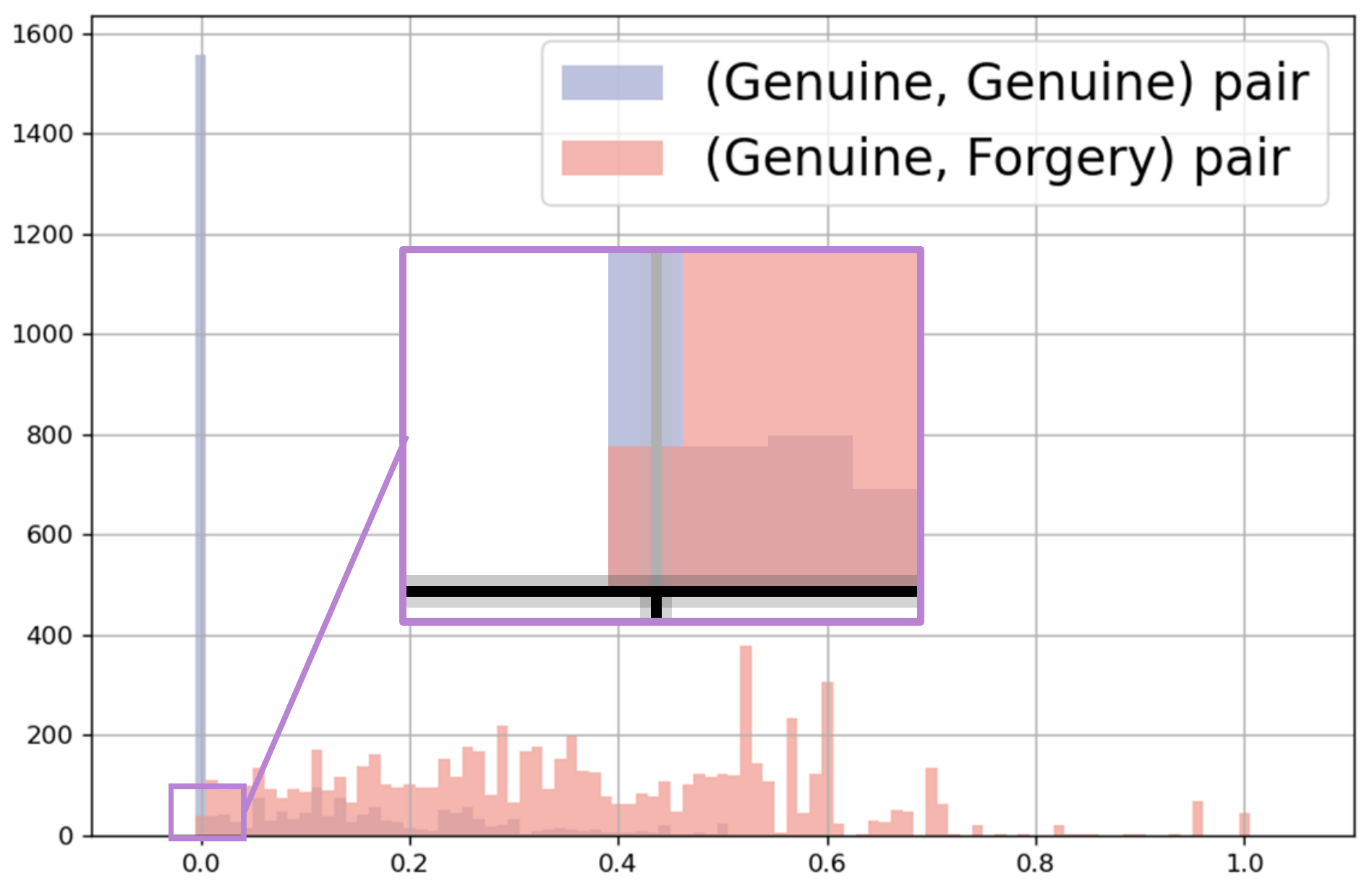}
\captionsetup{justification=centering}
\caption{Histogram of the ranking scores of SigNet with a zoomed view on BHSig-B}
\label{fig:bengalisighist}%文中引用该图片代号
\end{subfigure}
\centering
\begin{subfigure}{0.495\columnwidth}
\centering
\includegraphics[width=6cm]{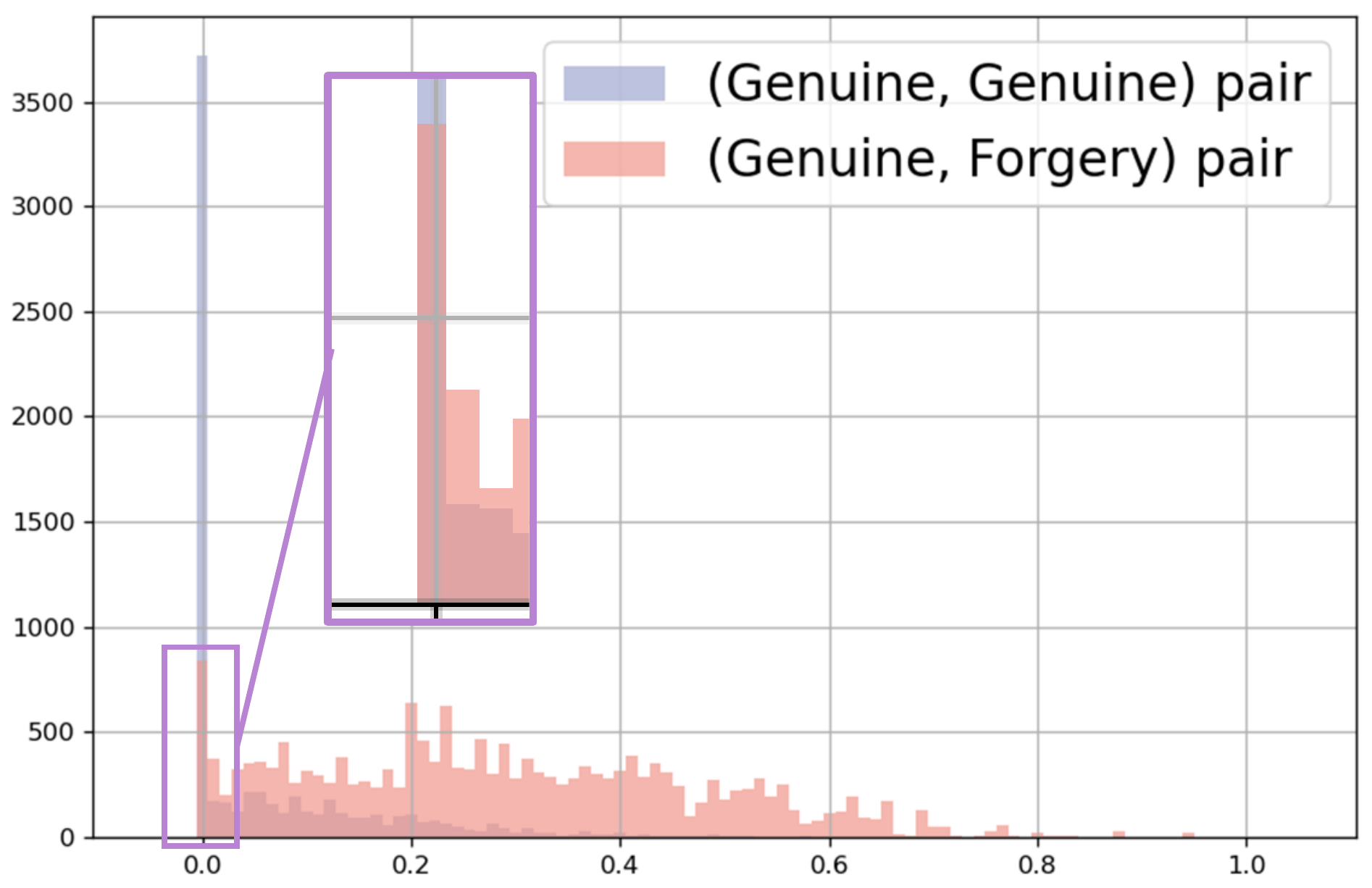}
\captionsetup{justification=centering}
\caption{Histogram of the ranking scores of SigNet with a zoomed view on BHSig-H}
\label{fig:hindisighist}%文中引用该图片代号
\end{subfigure}
\centering
\caption{(a) and (b)~PCA visualizations of feature distribution for the proposed method. The top-ranked negative and the absolute top positives are highlighted. (c)-(f)~Distributions of the ranking scores as histograms. The horizontal and vertical axes represent the ranking score and \#samples, respectively. }
\label{fig:hist}
\end{figure}
%%%%%%%%%%%%%%%%%%%%%%%%%%%%%%%%%%%%%%%

The quantitative evaluations of the proposed method and SigNet on two datasets are shown in Table~\ref{tab1}.
Remarkably, the proposed method achieved an overwhelming better performance on pos@top, while pos@top of SigNet is 0 for both datasets. This proves that the proposed method can reveal absolute top positive signature pairs (i.e., highly reliable signature pairs). Furthermore, the proposed method also outperformed the comparison method on all other evaluation criteria, accuracy and AUC, lower FAR, and FRR.\par

The ROC curves of the proposed method and SigNet on two datasets are shown in Fig.~\ref{fig:roc} respectively, each followed by a corresponding zoom view of the beginning part of ROC curves. As a more intuitive demonstration of Table~\ref{tab1}, other than the larger AUC, it is extremely obvious that the proposed method achieved higher pos@top, which is the True Positive Rate (TPR) for $x=0$.\par

Figs.~\ref{fig:hist}~(a) and (b) show the distributions of features from the proposed method mapped by Principal Component Analysis (PCA) on two datasets, colored by ranking scores (normalized within the range [0,1]). %It is worth mentioning that the best performing hyper-parameter $p$s were fixed by validation set respectively for each dataset, where $p$=4 for BHSig-B and $p$=16 for BHSig-H. 
From the visualizations, we can easily notice the absolute top positive sample pairs distinguished by the top-ranked negative. \par

Following the feature distributions, Figs.~\ref{fig:hist}~(c) and (d)~give a more intuitive representation of the ranking order for the learning top-rank pairs. These graphs include information of (1)~where did the top-ranked negative appear and (2)~how many (Genuine, Genuine) and (Genuine, Forgery)-pairs are scored to the same rank. It could be observed that the first negative appeared after a portion of positive from these two graphs. On the other hand, Figs.~\ref{fig:hist}~(e) and (f)  show the ranking conditions of SigNet. Since the first negative pair shows on the top of the ranking, this causes 0 pos@top for both datasets. \par
%By the way, the effectiveness of SigNet to minimize the distance between positive pairs is shown on graphs, which causes a concentration of positive pairs in $x=0$.

%%% fig Example
%%%%%%%%%%%%%%%%%%%%%%%%%%%%%%%%%%%%%%%%%%%
\setlength{\belowcaptionskip}{-0.5cm}
\begin{figure}[t!]
\begin{center}
\includegraphics[width=\columnwidth]{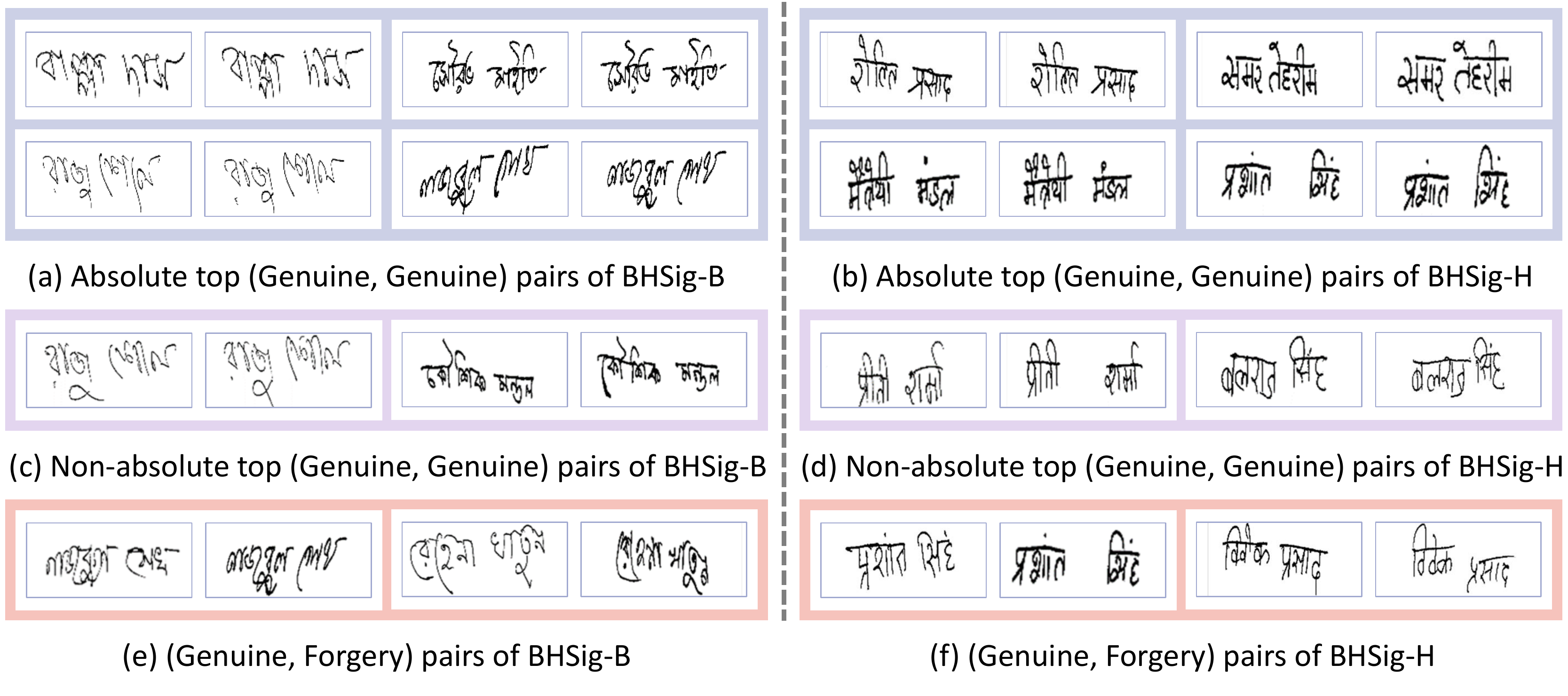}
\caption{Examples of (a)(b)~absolute top (Genuine, Genuine)-pairs, (c)(d)~non-absolute top (Genuine, Genuine)-pairs, and (e)(f)~(Genuine, Forgery)-pairs from BHSig-B and BHSig-H respectively.}
\label{fig:example}
\end{center}
\end{figure}
%%%%%%%%%%%%%%%%%%%%%%%%%%%%%%%%%%%%

As shown in Fig.~\ref{fig:example}, the absolute top (Genuine, Genuine)-pairs in (a) and (b) show great similarity to their counterparts. Especially, the consistency in their strokes and preference of oblique could be easily noticed even with the naked eye. On the other hand, both the non-absolute top (Genuine, Genuine)-pairs in (c) and (d) and (Genuine, Forgery)-pairs in (e) and (f) show less similarity to their corresponding signatures, no matter whether they are written by the same writer or not. Notwithstanding that they are all assigned with low ranking scores by learning top-rank pairs, such resemblance between two different classes could easily incur misclassification in conventional methods. Thus, our results shed light on the validity of the claimed effectiveness of the proposed method to maximize the pos@top.

\section{Conclusion}
As a critical application especially for formal scenarios like forensic handwriting analysis, signature verification played an important role since it has been proposed. In this work, we proposed a writer-independent signature verification model for learning top-rank pairs. What is novel and interesting for this model is that the optimization objective of top-rank learning is to maximize the pos@top, to say the highly reliable signature pairs in this case. This optimization goal has fulfilled the requirement of the intuitive need of signature verification tasks to acquire reliable genuine signatures, not only to naively classify positive from negative. Through two experiments on data set BHSig-B and BHSig-H, the effectiveness of pos@top maximization has been proved compared with a metric learning-based network, the SigNet. Besides, the performance of the proposed model on the AUC, accuracy, and other common evaluation criteria frequently used in signature verification shown encouraging results as well.

\newcommand{\BIBdecl}{\setlength{\itemsep}{0.2mm}}
\bibliographystyle{IEEEtran}
\bibliography{Topranksignature}

\end{document}